\documentclass{article}
\usepackage[preprint]{spconf,amsmath,graphicx}
\copyrightnotice{\copyright\ IEEE 2024}
\toappear{To appear in {\it Proc.\ ICASSP2024, April 14-19, 2024, Seoul, Korea}}
\usepackage{amsfonts}
\usepackage{hyperref}
\usepackage{bm}
\usepackage{float}
\usepackage{nicefrac}
\usepackage{support-caption}
\usepackage{subcaption}
\usepackage{multirow}
\usepackage{multicol}
\usepackage{makecell}
\usepackage{adjustbox}
\usepackage{booktabs}
\usepackage{tablefootnote}
\usepackage{xspace}

\def\resformer{\emph{ResidualTransformer}\xspace}

\title{ResidualTransformer: Residual Low-Rank Learning with Weight-Sharing for Transformer Layers}
\name{Yiming Wang \qquad Jinyu Li}
\address{
  Microsoft Corporation, Redmond, WA, USA \\
\small{\texttt{\{yimingwang,jinyli\}@microsoft.com}}
}
\begin{document}
\ninept
\maketitle
\begin{abstract}
Memory constraint of always-on devices is one of the major concerns when deploying speech processing models on these devices. While larger models trained with sufficiently large amount of data generally perform better, making them fit in the device memory is a demanding challenge. In this paper, we aim to reduce model size by reparameterizing model weights across Transformer encoder layers and assuming a special weight composition and structure. More specifically, inspired by ResNet \cite{he2016deep} and the more recent LoRA \cite{hu2022lora} work, we propose an approach named \resformer, where each weight matrix in a Transformer layer comprises 1) a shared full-rank component with its adjacent layers, and 2) a unique low-rank component to itself. The low-rank matrices only account for a small amount of model size increase. In addition, we add diagonal weight matrices to improve modeling capacity of the low-rank matrices. Experiments of our 10k-hour speech recognition and speech translation tasks show that the Transformer encoder size can be reduced by $\sim$3$\times$ with very slight performance degradation.

\end{abstract}
\section{Introduction}
\label{sec:intro}

Recent success on neural network based large models (usually at least hundreds of millions of parameters) has led to a series of breakthroughs in natural language and speech processing tasks \cite{devlin2019bert,brown2020language,radford2023robust,zhang2023google}. While there is evidence that these large models, if trained with sufficiently large amount of data, are of crucial importance for achieving state-of-the-art performance, they also require more run-time memory (or sophisticated distributed inference mechanism, e.g., \cite{du2022energonai}) and fast memory loading speed for model weights\footnote{We use ``weights'' and ``parameters'' interchangeably when referring to the variables, typically in the form of matrices, learned from model training.}.

On the other end of the research spectrum, sometimes people are more interested in ``small models'' when they are deploying AI models to always-on, embedded devices (e.g., smart phones, digital assistants, wearable devices, etc.). Such devices are usually equipped with limited computational resources, power source, and memory, requiring low peak memory usage during inference \cite{wang2021wake}. One solution to fitting a model to small memory is splitting the model into parts and continuously loading these partial models (e.g., layer by layer). However, it will incur increased power usage and inference latency \cite{hernandez2023sharing}. In such scenarios, it would be desirable to have small models that fit in the memory and still maintain good performance.

In this paper, we aim to reduce the model size by sharing most of model weights across layers and assuming a special composition and structure of the weight matrices. During model inference, the shared weights are loaded into memory once, saving time cost spent on model loading for devices with limited memory. The approach was motivated by observations from our preliminary experiments where model weights are shared across consecutive Transformer layers in the encoder network. It suggests that parameters may not differ too much across a few layers.

We focus on applying our method to the Transformer layers \cite{vaswani2017attention}, as the Transformer architecture has been widely adopted as a building block in various natural language and speech processing tasks \cite[\emph{inter alia}]{dai2019transformer,brown2020language,dong2018speech,zhang2020transformer,wang2021wake2}, and usually it accounts for a significant portion of the total number of model parameters. It should be noted that our method can also be applied to weight matrices in other types of neural network layers (e.g., LSTM, embeddings, convolution, etc.), which will be our future work.

\section{Related Work}
\label{sec:related}

Many efforts have been made in order to deploy models on memory-limited devices. These methods include knowledge distillation \cite{hinton2015distilling,pang2018compression}, low-rank factorization \cite{xue2014singular,winata2020lightweight}, low-bit quantization \cite{han2016deep,ding2022bit}, and model pruning \cite{han2015learning,zhu2018prune}. They either require special hardware support which is not always available on the devices (quantization), have considerable performance loss (low-rank factorization, knowledge distillation), or are not very efficient in terms of training (knowledge distillation) or tuning (model pruning). Our method is orthogonal to most of these approaches (e.g., quantization, model pruning, and knowledge distillation), has no special hardware requirements, and leads to very slight performance loss.

The idea of reusing Transformer layer was proposed in \cite{dehghani2019universal} for natural language processing tasks with a different motivation: they regard the repeated applications of the network function as a complementary way of introducing a recurrent inductive bias to Transformers, and it was observed that their method outperforms the vanilla Transformer on several tasks. \cite{lan2020albert} adopted the cross-layer weight-sharing in the pretraining/fine-tuning setting to improve model scalability. \cite{gao2021extremely} extends it to the speech recognition task, working on relatively small-scale datasets (100--1k hours). The results with weight-sharing in their settings are comparable or even outperforms the unsharing baselines, which is not the case in our 10k-hour setups. \cite{hernandez2023sharing} explored sharing different parts of Conformer \cite{gulati2020conformer} on different levels of granularity with the model size hard-constrained by memory.

Low-rank factorization was applied to the parameters of Transformers in \cite{winata2020lightweight}. WER improvement was observed with a properly chosen rank on their 150-hour datasets. However, our preliminary experiments indicate that it causes degradation on large-scale tasks (e.g., 10k hours). Also, \cite{lv2023lightformer} demonstrated that it is less parameter-efficient than weight sharing across layers. LoRA \cite{hu2022lora} was recently proposed to adapt a large model (e.g., GPT-3 \cite{brown2020language}), well-(pre-)trained on general domains, to a specific domain or task. Instead of fine-tuning all the parameters which is time and resource consuming, LoRA freezes all those pretrained parameters, and injects very small amount of trainable low-rank decomposed weights to each layer for fine-tuning. Our approach was inspired by LoRA, but with a different goal: we would like the low-rank weights to compensate for the performance loss resulted from weight sharing, without increasing the model size significantly. Because of the nature of this different problem, we do not freeze the shared weights during training.

Our method is also inspired by ResNet \cite{he2016deep}, where a residual block is to learn the residual function after subtracting the identity mapping. In our case, low-rank ``residual weights'' of Transformer layers are added and trained together with the shared weights, which is different from simply replacing the full-rank weights described in \cite{winata2020lightweight,hernandez2023sharing}. Therefore, we name our method \resformer.

\section{Methods}
\label{sec:methods}

\subsection{Transformers}
\label{sec:Transformers}
Transformers \cite{vaswani2017attention}, characterized by its self-attention structure and parallelized processing over sequences, has gained popularity in both natural language and speech communities for its capability of directly modeling context dependency for sequence data. The Transformer architecture is now a building block of various models in sequence modeling tasks. For speech recognition and translation being focused on in this paper, Transformers layers, which usually contribute a lot (around 70\% in our cases) to the model size, are part of the encoder network. Our method is applied to all the projection weights in the Transformer layers, including the linear projections for query, key and value in the self-attention module, the linear post-projection after self-attention, and those in the feed-forward module. Let $x \in \mathbb{R}^M$ be the input and $y \in \mathbb{R}^N$ the output, these projections can be formulated as:

\begin{equation}
    y=W^\top x+b
\end{equation}
where the projection matrix $W\in \mathbb{R}^{M \times N}$ and the bias vector $b\in \mathbb{R}^N$ are trainable weights.

\subsection{Weight Sharing across Layers}
\label{sec:weight_sharing}

Assume we have an $L$-layer Transformer encoder, with a weight matrix $W_l \in \mathbb{R}^{M \times N}$ associated with the $l$-th layer \footnote{The bias vector $b_l$ of the weights is treated similarly regarding weight sharing.}. Normally, $W_l$ are distinct to each other, as each layer has their own parameters. We denote these distinct matrices as $U_l \in \mathbb{R}^{M \times N}$, such that $W_l=U_l$ for $l \in [0,L-1]$. In this case, the total number of the parameters is $L \times M \times N$.

Due to memory constraint, we are going to reduce the model size by dividing the $L$ layers into groups, and making all the layers within the same group to share common weight matrices (while Dropout and LayerNorm modules still operate independently). Specifically, every consecutive $K$ layers are grouped together, so that $W_0=U_0,W_1=U_0,\ldots,W_{K-1}=U_0,W_K=U_1,\ldots$ (or more generally, with distinct matrices $\{U_i|i=0,\ldots,\lceil L/K \rceil-1\}$, we have $W_l=U_{\lceil l/K \rceil}$). Consequently, a weight matrix is reused $K$ times, and the total number of the parameters is $\lceil L/K \rceil \times M \times N$, typically $K$ times smaller than the original network. During inference, weights can be sequentially loaded into memory once every $K$ layers in the forward pass of the network, saving the time cost of loading the model by the factor of $K$.

\subsection{Adding Residual Matrices to Shared Weights}
\label{sec:resformer}

Our preliminary speech recognition experiments (see Section \ref{sec:results:weight_sharing}) reveal that, with a 18-layer encoder, the model with $K=3$ only performs 4\% worse relatively than the baseline (i.e., $K=1$), and the gap quickly becomes larger as $K$ increases, e.g., the model with $K=9$ is 17\% worse. This may suggest that, if we consider the model training as an optimization problem, there exists a close-to-optimal solution where the layer parameters does not differ too much across a few consecutive layers. To compensate for performance loss due to the weight sharing strategy across layers within the same group,  we add a few more parameters to each layer, so that a layer can still play as a unique transforming function. Inspired by ResNet \cite{he2016deep}, where the parameters in a residual block is supposed to learn the residual function in addition to the identity mapping, we add a ``residual'' weight matrix $\Delta W_l \in \mathbb{R}^{M \times N}$ to the $l$-th layer to learn a layer-specific residual function:

\begin{equation}
    y=(W_l^\top +\Delta W_l^\top) x+b_l
\end{equation}
Ideally $\Delta W_l$ should have much smaller number of parameters compared to $W_l$. A good candidate of it could be low-rank decomposed matrices of the form $\Delta W=AB$ where $A \in \mathbb{R}^{M \times R}$, $B \in \mathbb{R}^{R \times N}$ and $R \ll \min(M,N)$. At first glance it looks very similar to LoRA \cite{hu2022lora} where such low-rank decomposed weights are added to a well-trained large model to adapt the model to another particular domain or task, by only fine-tuning the low-rank weights on the target domain/task. Our approach is inspired by LoRA. However, it faces a different problem and has a different purpose from LoRA: the model (before adding the low-rank matrices) is built under resource constraints, thus is not a full-fledged one (due to weight sharing across layers); and we are not doing adaptation. Therefore, we need to update all the parameters, including $W_l$, during training to achieve the best performance.

As in \cite{zhao2017extended}, we also add a diagonal matrix\footnote{More strictly, we are referring to \emph{rectangular diagonal matrix}, which is an M-by-N matrix with all the entries not indexed by $(i,i)$ being zero.} $D_l \in \mathbb{R}^{M \times N}$ to $\Delta W_l$, making $\Delta W_l$ full rank without introducing a significant number of parameters:

\begin{equation}
    \Delta W_l=A_lB_l+D_l
\end{equation}

\begin{figure}[ht]
\centering
\includegraphics[width=0.5\textwidth]{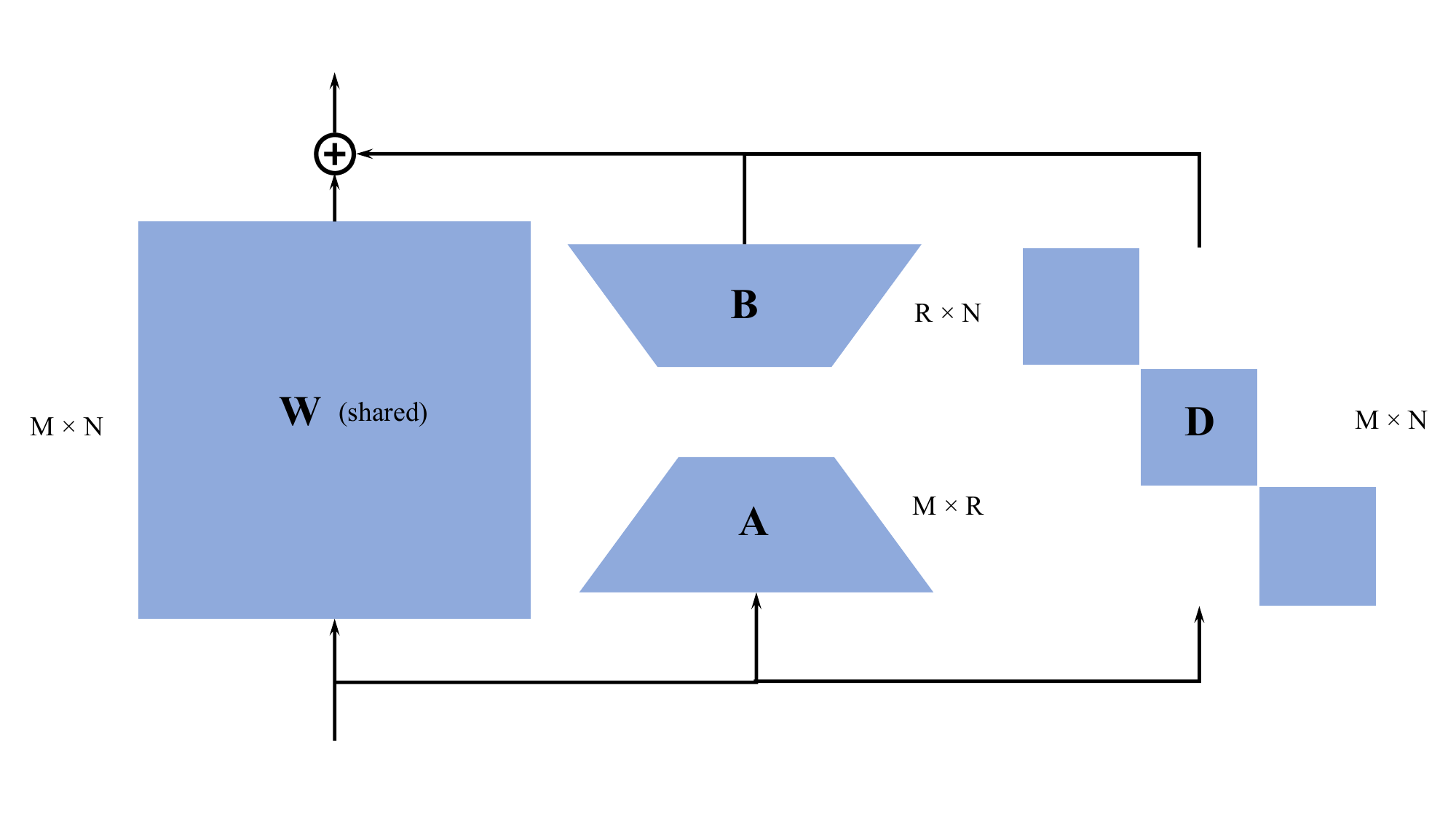}
\caption{The proposed weight structure for each weight matrix in a Transformer layer. The layer index $l$ is omitted in the notations. $W$ is a full-rank matrix being shared with other layers, $A$ and $B$ are two low-rank decomposed matrices, and $D$ is a diagonal matrix.}
\label{fig:model}
\end{figure}

Fig. \ref{fig:model} illustrates the proposed weight matrix structure. $A$, $B$ and $D$ together are designated as \emph{residual matrices} for the shared one $W$, so that a unique function will be learned for each Transformer layer despite the presence of weight sharing. We name this structure as \resformer.

\section{Experiments}
\label{sec:expr}

\vspace{-2mm}
\subsection{Datasets}
\label{sec:datasets}

Both the speech recognition (ASR) and translation (ST) task use 10k hours of our in-house anonymized English audio data for training. For ASR, the original English transcriptions are tokenized into sentence pieces \cite{kudo2018sentencepiece}. For ST, we leverage Microsoft cognitive translation service to translate the original English (EN) transcriptions into Chinese (ZH), and use those translated text as training targets after tokenization by characters.

The test set for ASR covers various application scenarios, such as Cortana, far-field speech and call center, consisting of 1.8M words in total. The test sets for EN-ZH ST include the publicly available CoVoST 2 \cite{wang2021covost} (a total of 61k characters in reference, denoted as \textit{C}), MSLT v1.1 \cite{federmann2016microsoft} dev and test sets (a total of 92k characters in reference, denoted as \textit{M}), and our internal data covering four scenarios of call center, conversation, dictation, and meetings (a total of 317k characters in reference, denoted as \textit{I}). When reporting BLEU scores on a data set, numbers are averaged over its subsets. Note that our ST training set does not contain any in-domain data of the public test sets, i.e., we do not use any training data from CoVoST 2 or MSLT.

\vspace{-2mm}
\subsection{Models}
\label{sec:models}

The Transformer-Transducer model \cite{graves2012sequence,zhang2020transformer} with chunk-wise masks for streaming \cite{chen2021developing} tasks is used to evaluate our method for ASR and ST. The encoder consists of 2 2D convolutions with a total of 4$\times$ downsampling, 18 Transformer layers with relative positional embeddings, where each layer includes a 512-dim multi-head attention with 8 heads and a 2048-dim feed-forward layer. The masks are configured such that the chunk size is 160ms, and for each chunk the look-back history is 288ms and the look-ahead latency is 160ms. The prediction network is a 2-layer LSTM network with the hidden size of 1024 and the embedding size of 320. The joint network projects the output of the encoder and prediction network into a 512-dim space. The input to the network is 80-dim FBANK features with a 10ms stride, normalized by the global mean and variance computed on the training set. Depending on the task, the vocabulary size is 4,002 (for ASR) or 11,109 (for ST). The total size of the baseline model is 78.0M (for ASR) or 83.9M (for ST), 56.7M of which are parameters of the Transformer layers. In this paper, we only investigate applying our method to Transformers, and leave applications to other types of network blocks for future work.

\vspace{-2mm}
\subsection{Training}
\label{sec:training}
Baseline models, and models with weight sharing but without residual weights, are trained from scratch (i.e., parameters are randomly initialized) for 350k/200k \footnote{No further improvement is observed if trained with more steps.} steps with Adam optimizer and linearly decaying learning rate on 16/32 GPUs for ASR/ST. When it comes to models with \resformer that include residual matrices, all but the residual weights are initialized with the parameters from their weight-sharing-only counterparts, and then trained for the same number of steps as their respective baseline does. Compared to training from scratch, we found that initialization in this way empirically benefits the final model performance.

\vspace{-2mm}
\subsection{Results}
\label{sec:results}

\subsubsection{Weight Sharing}
\label{sec:results:weight_sharing}

We first apply the weight sharing described in Section \ref{sec:weight_sharing} to the 18-layer Transformer encoder (i.e., $L=18$), and set $K$ to 1 (baseline), 3, 6, 9, and 18, respectively. We also train models without weight sharing (i.e., $K=1$) with $L=$ 6, 3, 2, and 1, respectively, for comparisons between models of the same size. The ASR results are shown in Table \ref{tab:weight_sharing_asr}. Note that models in the same column contain about the same number of parameters\footnote{There are two LayerNorm modules in a Transformer layer whose trainable parameters are not shared. But the number of these parameters is negligible. So we still consider the two models of the same size.}. We have two observations. 1) Having the same model size, the model with weight sharing performs much better than the one without weight sharing (thanks to repeatedly applying the same layer function $K$ times), and the gap becomes larger when the model size is smaller---WER is almost halved in the last column. 2) With weight sharing, WER degrades faster as $K$ increases. For example, when $K$ increases from 1 to 3, WER is only hurt by 4.0\% relative (13.28$\rightarrow$13.81). But when $K$ increases from 6 to 18, the relative change is as high as 18.3\% (14.59$\rightarrow$17.26). The ST results in Table \ref{tab:weight_sharing_st} demonstrate a similar trend. These imply that: 1) weight sharing is an effective way of improving model capacity without increasing the model size; and 2) when sharing weights, it would be better to share with adjacent layers than with those far away. The parameters of a distant layer are probably quite different, and sharing weights with it will limit the model's flexibility of learning good speech representations through a stack of Transformers.

\begin{table}[ht]
\caption{Comparisons of ASR models with or without weight sharing. Models in the same column are about of the same size.}
\vspace{-3mm}
\begin{center}
\begin{adjustbox}{max width=\linewidth}
\begin{tabular}{lccccc}
\toprule
no weight sharing: $K=1$, $L=$ &  & 6 & 3 & 2 & 1 \\
\cmidrule(lr){3-6}
WER(\%)$\downarrow$           &  & 16.25 & 20.11 & 23.60 & 33.23 \\ 
\midrule
weight sharing: $L=18$, $K=$ & 1 (baseline) & 3 & 6 & 9 & 18 \\
\cmidrule(lr){2-6}
WER(\%)$\downarrow$ & 13.28 & \textbf{13.81} & 14.59 & 15.49 & 17.26 \\
\bottomrule
\end{tabular}
\end{adjustbox}
\end{center}
\label{tab:weight_sharing_asr}
\end{table}

\vspace{-4.5mm}
\begin{table}[ht]
\caption{Comparisons of ST models with or without weight sharing. Models in the same column are about of the same size. BLEU is reported on three sets individually.}
\vspace{-3mm}
\LARGE
\begin{center}
\begin{adjustbox}{max width=\linewidth}
\begin{tabular}{lccccc}
\toprule
no weight sharing: $K=1$, $L=$ &  & 6 & 3 & 2 & 1 \\
\cmidrule(lr){3-6}
BLEU(\%)$\uparrow$  \quad \textit{C}\(|\)\textit{M}\(|\)\textit{I}  &  & 22.3\(|\)28.2\(|\)26.0 & 16.1\(|\)23.2\(|\)22.6 & 12.9\(|\)20.1\(|\)20.2 & 7.3\(|\)14.2\(|\)15.9 \\ 
\midrule
weight sharing: $L=18$, $K=$ & 1 (baseline) & 3 & 6 & 9 & 18 \\
\cmidrule(lr){2-6}
BLEU(\%)$\uparrow$ \quad \textit{C}\(|\)\textit{M}\(|\)\textit{I} & 29.8\(|\)32.4\(|\)29.4 & \textbf{26.7}\(|\)\textbf{31.3}\(|\)\textbf{28.3} & 24.7\(|\)29.9\(|\)27.4 & 23.6\(|\)29.3\(|\)26.5 & 19.1\(|\)25.6\(|\)24.5 \\
\bottomrule
\end{tabular}
\end{adjustbox}
\end{center}
\label{tab:weight_sharing_st}
\end{table}

\vspace{-8mm}
\subsubsection{Residual Weights}
\label{sec:results:residual_weight}

Next, we add low-rank and diagonal residual weights to the shared weights. We set $R=16$ so that, for example, for a weight matrix in the feed-forward module $W^\mathrm{F} \in \mathbb{R}^{512 \times 2048}$, a pair of low-rank decomposed matrices $A^\mathrm{F} \in \mathbb{R}^{512 \times 16}$ and $B^\mathrm{F} \in \mathbb{R}^{16 \times 2048}$ along with a diagonal matrix $D^\mathrm{F} \in \mathbb{R}^{512 \times 2048}$ are added to each Transformer layer, summing up to the model size increase of 2.7M. The effect of adding such residual weights under different values of $K$ is shown in Table \ref{tab:residual_weights_asr} (for ASR) and Table \ref{tab:residual_weights_st} (for ST). Apparently, adding residual weights consistently reduces the relative gap between weight sharing and the baseline. For example, when $K=3$, the gap is reduced from 4.0\% to 1.8\% in ASR, or from 3.4--10.4\% to 0.3--3.4\% in ST, i.e., the gap is reduced by more than 55\% relative, and the remaining gap is quite small given such a large reduction in model size. However, the extent to which the gap is reduced becomes smaller when $K$ is larger. For instance, when $K=18$, the relative gap is reduced from 30.0\% to 21.2\% in ASR, or from 16.7--35.9\% to 13.3--27.2\% in ST, a reduction by only 20+\% relative. This suggest that, if too many layers are grouped together (i.e., $K$ is too large) for weight sharing, adding the residual weights to each layer is still unable to compensate for most of the gap. This can be attributed to the underlying structure of the residual weights. Although we make it full-rank by adding a diagonal matrix, the residual weights are still not as powerful as full matrices in the baseline that are unique to each layer.

The comparison between ``weight sharing with $K=6$'' and ``+ residual weights with $K=9$'' also caught our attention. These two settings have the similar number of parameters (9.5M vs. 9.0M) and their performance is close (14.59 vs. 14.62 in WER, 24.7\(|\)29.9\(|\)27.4 vs. 25.1\(|\)30.2\(|\)27.3 in BLEU). A natural question arises: is the gain obtained from adding residual weights simply because of the model size increase? To answer this question, we make the weights of the last layer\footnote{we also tried to make the middle or 2nd layer unique, and the results are similar to or worse than doing so to the top layer.} unique in the setting ``weight sharing with $K=3$'', so that the number of parameters increases from 18.9M to 22.0M. We trained this model for ASR and the WER of such a model is 13.73, worse than 13.52 from the setting ``+ residual weights with $K=3$'' (21.6M params.). Similarly, if the 4 sets of distinct layer weights spread roughly evenly across the 18 layers (12.6M params.), the WER is 14.28, still worse than 14.12 from the setting ``+ residual weights with $K=6$'' (12.2M params.). These additional experiments provide more evidence to support our previous speculation that when $K$ is not that large (e.g., $<=$6), using residual weights is more parameter-efficient than just adding distinct layer weights.

\vspace{-2mm}
\begin{table}[ht]
\caption{Effect of adding residual weights for ASR under different values of weight-sharing factor $K$ (with rank $R=16$).}
\vspace{-3mm}
\begin{center}
\begin{adjustbox}{max width=\linewidth}
\begin{tabular}{llccccc}
\toprule
\multicolumn{2}{c}{$K$} & 1 (baseline) & 3 & 6 & 9 & 18 \\
\midrule
\multirow{3}{*}{weight sharing} & \# params. in Trans.  & 56.7M & 18.9M & 9.5M & 6.3M & 3.2M \\
& WER(\%)$\downarrow$ & 13.28 & 13.81 & 14.59 & 15.49 & 17.26 \\
& rel. gap from baseline (\%) & & 4.0 & 9.9 & 16.6 & 30.0 \\
\midrule
\multirow{3}{*}{\quad + residual weights} & \# params. in Trans.  & N/A & 21.6M & 12.2M & 9.0M & 5.9M \\
& WER(\%)$\downarrow$    &    N/A  & \textbf{13.52} & 14.12 & 14.62 & 16.09 \\
& rel. gap from baseline (\%) & & 1.8 & 6.3 & 10.1 & 21.2 \\
\bottomrule
\end{tabular}
\end{adjustbox}
\end{center}
\label{tab:residual_weights_asr}
\end{table}

\vspace{-4.5mm}
\begin{table}[ht]
\caption{Effect of adding residual weights for ST under different values of weight-sharing factor $K$ (with rank $R=16$). BLEU is reported on three sets individually.}
\vspace{-3mm}
\LARGE
\begin{center}
\begin{adjustbox}{max width=\linewidth}
\begin{tabular}{llccccc}
\toprule
\multicolumn{2}{c}{$K$} & 1 (baseline) & 3 & 6 & 9 & 18 \\
\midrule
\multirow{3}{*}{weight sharing} & \# params. in Trans.  & 56.7M & 18.9M & 9.5M & 6.3M & 3.2M \\
& BLEU(\%)$\uparrow$~~\textit{C}\(|\)\textit{M}\(|\)\textit{I} & 29.8\(|\)32.4\(|\)29.4 & 26.7\(|\)31.3\(|\)28.3 & 24.7\(|\)29.9\(|\)27.4 & 23.6\(|\)29.3\(|\)26.5 & 19.1\(|\)25.6\(|\)24.5 \\
& rel. gap from baseline (\%) & & 10.4\(|\)3.4\(|\)3.7 & 17.1\(|\)7.7\(|\)6.8 & 20.8\(|\)9.6\(|\)9.9 & 35.9\(|\)21.0\(|\)16.7 \\
\midrule
\multirow{3}{*}{\quad + residual weights} & \# params. in Trans.  & N/A & 21.6M & 12.2M & 9.0M & 5.9M \\
& BLEU(\%)$\uparrow$~~\textit{C}\(|\)\textit{M}\(|\)\textit{I} &    N/A  & \textbf{28.8}\(|\)\textbf{32.3}\(|\)\textbf{29.0} & 26.5\(|\)30.9\(|\)28.3 & 25.1\(|\)30.2\(|\)27.3 & 21.7\(|\)27.7\(|\)25.5 \\
& rel. gap from baseline (\%) & & 3.4\(|\)0.3\(|\)1.4 & 11.1\(|\)4.6\(|\)3.7 & 15.8\(|\)6.8\(|\)7.1 & 27.2\(|\)14.5\(|\)13.3 \\
\bottomrule
\end{tabular}
\end{adjustbox}
\end{center}
\label{tab:residual_weights_st}
\end{table}

\vspace{-5mm}
\subsubsection{Rank of Matrices}
\label{sec:results:rank}

We are now going to see how the model performance will be affected with different values of rank $R$. Due to the memory constraint, we do not experiment with settings where $R>16$, and instead focus on smaller values of $R$, i.e., 16, 8, 4, 2, and 1. We show the results of ASR with $K=$ 3 or 9 in Table \ref{tab:rank_asr}. It can be seen that, with $K=3$, decreasing the value of $R$ does not have significant impact on WER until $R$ reaches 1. Even when $R=1$, the WER (=13.59) is still better than 13.73 which is from the aforementioned configuration where weights in the last layer are unique and the model size is larger (19.1M vs. 22.0M). With $K=9$, however, we do observe that a smaller $R$ hurts the performance a little bit, since given a larger sharing factor $K$, it may need a larger matrix rank for layers to distinguish themselves among others within the same weight-sharing group. Due to space limitations we do not tabulate the ST results which have a similar tendency, and only report BLEU when $K=3$ and $R=2$, which is, 28.8\(|\)32.3\(|\)29.2.

\vspace{-2mm}
\begin{table}[ht]
\caption{Effect of different values of rank $R$ of residual weights for ASR with the weight-sharing factor $K=$ 3 or 9.}
\vspace{-3mm}
\LARGE
\begin{center}
\begin{adjustbox}{max width=\linewidth}
\begin{tabular}{lccccc}
\toprule
$R$ & 16 & 8 & 4 & 2 & 1\\
\midrule
$K=3$ \\
 \# params. in Trans. (residual weights)  & 21.6M (2.7M) & 20.3M (1.4M) & 19.6M (0.7M) & 19.3 (0.4M) & 19.1M (0.2M) \\
WER(\%)$\downarrow$ & 13.52 & \textbf{13.51} & 13.54 & 13.53 & 13.59 \\
rel. gap from baseline (\%) & 1.8 & 1.7 & 2.0 & 1.9 & 2.3 \\
\midrule
$K=9$ \\
\# params. in Trans. (residual weights)  & 9.0M (2.7M) & 7.7M (1.4M) & 7.0M (0.7M) & 6.7 (0.4M) & 6.5M (0.2M) \\
WER(\%)$\downarrow$ & \textbf{14.62} & 14.73 & 14.82 & 14.88 & 14.95 \\
rel. gap from baseline (\%) & 10.1 & 10.9 & 11.60 & 12.0 & 12.6 \\
\bottomrule
\end{tabular}
\end{adjustbox}
\end{center}
\label{tab:rank_asr}
\end{table}

\vspace{-5mm}
Overall, if we look at the configuration ``weight sharing + residual weights with $K=3$ and $R=2$'', the number of parameters in Transformers is 34.0\% of that in the baseline, with only slight performance decline in ASR (1.9\%) and ST (3.4\(|\)0.3\(|\)0.7\%).

\vspace{-1mm}
\subsection{Ablation Study: Diagonal Matrices}
\label{sec:results:ablation}
 We wonder how the diagonal matrices contribute to the model performance. As an ablation study, we remove diagonal matrices from residual weights, and report the WER changes in ASR under several configurations in Table \ref{tab:diagonal_asr}. Generally, adding diagonal matrices has quite small impact on the results. Its effect is more prominent when $K$ is larger (D1 vs. D2), or $R$ is smaller (D3 vs. D4), corresponding to the cases where the rest of the model is relatively weak (due to either with more weight sharing, or with a lower rank of residual weights). The changes of BLEU in ST are similar. For example, when $K=3$ and $R=16$, BLEU is changed from 28.8\(|\)32.3\(|\)29.0 to 28.5\(|\)31.9\(|\)29.1; and when $K=3$ and $R=2$, BLEU is changed from 28.8\(|\)32.3\(|\)29.2 to 28.9\(|\)31.8\(|\)29.0. Since the increase of the model size is negligible, we keep the diagonal matrices as part of the residual weights.

\begin{table}[ht]
\caption{Effect of diagonal matrices in residual weights for ASR.}
\vspace{-3mm}
\begin{center}
\begin{adjustbox}{max width=\linewidth}
\begin{tabular}{lcccc}
\toprule
& $K$ & $R$ &  WER(\%)$\downarrow$ w/ diagonal & WER(\%)$\downarrow$ w/o diagonal \\
\midrule
D1 & 3 & 16 & 13.52 & 13.51 \\
D2 & 9 & 16 & 14.62 & 14.65 \\
D3 & 3 & 8 & 13.51 & 13.52 \\
D4 & 3 & 2 & 13.53 & 13.57\\
\bottomrule
\end{tabular}
\end{adjustbox}
\end{center}
\label{tab:diagonal_asr}
\end{table}

\vspace{-7.5mm}
\section{Conclusion}
\label{sec:conclusion}

We propose an approach to reduce the model size and memory loading cost of Transformer layers without significant performance loss for memory-constrained devices: each weight matrix in a layer consists of a full-rank component being shared with its adjacent layers, and a distinct low-rank component plus a diagonal matrix not being shared with others. The low-rank and diagonal matrices only account for a small amount of model size increase but can largely improve the model capacity. Experiments on our 10k-hour speech recognition and translation tasks show that the size of Transformers can be reduced by almost 3$\times$ while performance deterioration is quite small.

\vfill\pagebreak

\bibliographystyle{IEEEbib}
\fontsize{8.9}{10.7}\selectfont
\bibliography{strings,refs}

\end{document}